\documentclass[10pt,twocolumn,letterpaper]{article}

\usepackage{cvpr}
\usepackage{times}
\usepackage{epsfig}
\usepackage{graphicx}
\usepackage{amsmath}
\usepackage{amssymb}

\usepackage{bm}
\usepackage{subfigure}
\usepackage{multirow}
\usepackage{array}
\usepackage{authblk}
\usepackage{makecell}

\usepackage[pagebackref=true,breaklinks=true,letterpaper=true,colorlinks,bookmarks=false]{hyperref}

\cvprfinalcopy % *** Uncomment this line for the final submission

\def\cvprPaperID{842} % *** Enter the CVPR Paper ID here

\ifcvprfinal\fi

\begin{document}

%%%%%%%%% TITLE
\title{AttnGAN: Fine-Grained Text to Image Generation \\with Attentional Generative Adversarial Networks}

\author[1]{Tao Xu\thanks{ work was performed when was an intern with Microsoft Research}} 
\author[2]{\;Pengchuan Zhang} 
\author[2]{\;Qiuyuan Huang} 
\author[3]{\\Han Zhang}
\author[4]{\;Zhe Gan}
\author[1]{\;Xiaolei Huang}
\author[2]{\;Xiaodong He}

\affil[ ]{$^{1}$Lehigh University  \;  $^{2}$Microsoft Research \; $^{3}$Rutgers University \;  $^{4}$Duke University}
\affil[ ]{\tt\small {\{tax313,~xih206\}@lehigh.edu, \{penzhan,~qihua,~xiaohe\}@microsoft.com}}
\affil[ ]{\tt\small {han.zhang@cs.rutgers.edu, zhe.gan@duke.edu}}
\renewcommand\Authands{, } 

\maketitle
% \thispagestyle{empty}

%%%%%%%%% ABSTRACT
\begin{abstract}
In this paper, we propose an Attentional Generative Adversarial Network (AttnGAN) that allows attention-driven, multi-stage refinement for fine-grained text-to-image generation.  With a novel attentional generative network, the AttnGAN can synthesize fine-grained details at different sub-regions of the image by paying attentions to the relevant words in the natural language description. In addition, a deep attentional multimodal similarity model is proposed to compute a fine-grained image-text matching loss for training the generator. The proposed AttnGAN significantly outperforms the previous state of the art, boosting the best reported inception score by 14.14\% on the CUB dataset and 170.25\% on the more challenging COCO dataset. A detailed analysis is also performed by visualizing the attention layers of the AttnGAN. It for the first time shows that the layered attentional GAN is able to automatically select the condition at the word level for generating different parts of the image. 
\end{abstract}

%%%%%%%%% BODY TEXT

\vspace{-5pt}
\section{Introduction}
\vspace{-5pt}
Automatically generating images according to natural language descriptions is a fundamental problem in many applications, such as art generation and computer-aided design. It also drives research progress in multimodal learning and inference across vision and language, which is one of the most active research areas in recent years~\cite{reed2016generative,reed2016learning,Han16stackgan,reed2016cvpr,FangGISDDGHMPZZ15,XuBKCCSZB15,SCN_CVPR2017, AgrawalLAMZPB17,YangHGDS16}

Most recently proposed text-to-image synthesis methods are based on Generative Adversarial Networks (GANs)~\cite{goodfellow2014generative}. A commonly used approach is to encode the whole text description into a global sentence vector as the condition for GAN-based image generation~\cite{reed2016generative,reed2016learning,Han16stackgan,Han17stackgan2}. Although impressive results have been presented, conditioning GAN only on the global sentence vector lacks important fine-grained information at the word level, and prevents the generation of high quality images. This problem becomes even more severe when generating complex scenes such as those in the COCO dataset~\cite{LinMBHPRDZ14}.

\begin{figure}[t]
 \small
 \centering
 \begin{tabular}{@{\hspace{0mm}}p{0.095\textwidth}@{\hspace{1mm}}p{0.095\textwidth}@{\hspace{1mm}}p{0.095\textwidth}@{\hspace{1mm}}p{0.095\textwidth}@{\hspace{1mm}}p{0.095\textwidth}@{\hspace{0mm}}}
  \multicolumn{5}{c@{\hspace{1mm}}}{\shortstack{\\ {\color{blue} this bird} is  {\color{red} red} with  {\color{blue} white} and has {\color{red} a very short beak}}} 
  \\
 \multicolumn{5}{l@{\hspace{1mm}}}{\includegraphics[width=0.316\linewidth]{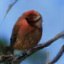} \includegraphics[width=0.316\linewidth]{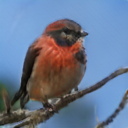} \includegraphics[width=0.316\linewidth]{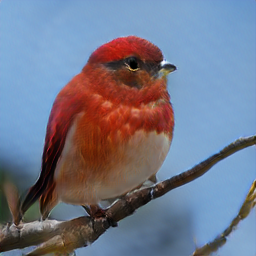}}
  \\
  \shortstack{\color{red}~~~~10:short} &\shortstack{\color{red}3:red} &\shortstack{\color{red}11:beak} &\shortstack{\color{red}9:very} &\shortstack{\color{red}8:a}
  \\
 \multicolumn{5}{l@{\hspace{1mm}}}{\includegraphics[width=0.967\linewidth]{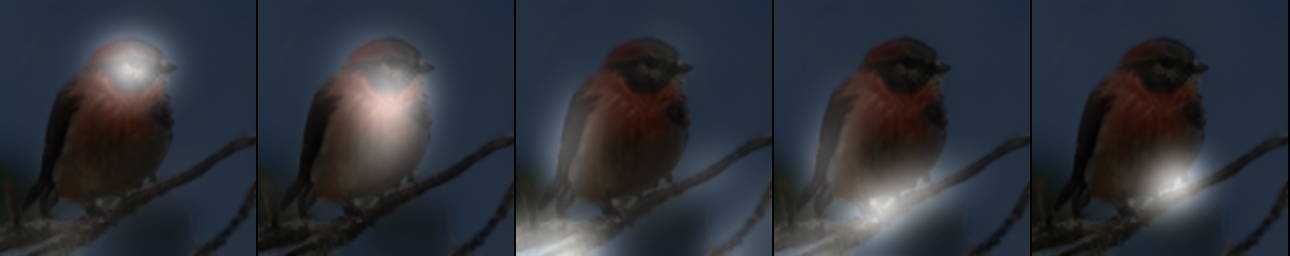}} 
  \\
  \shortstack{\color{blue}~~~~3:red} &\shortstack{\color{blue}5:white} &\shortstack{\color{blue}1:bird} &\shortstack{\color{blue}10:short} &\shortstack{\color{blue}0:this}
  \\
  \multicolumn{5}{l@{\hspace{1mm}}}{\includegraphics[width=0.967\linewidth]{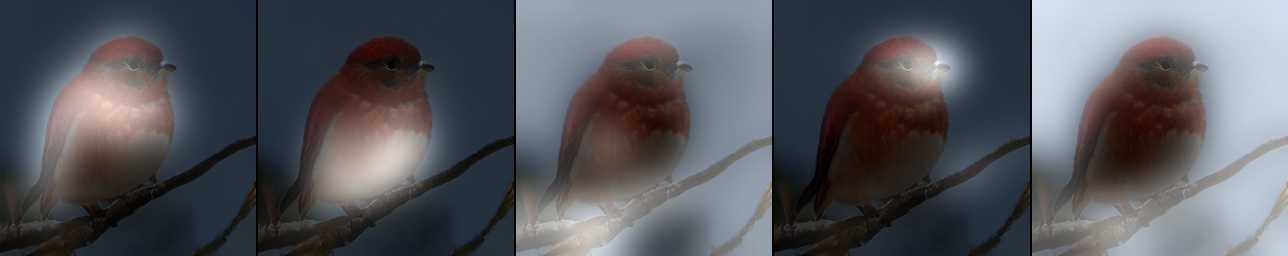}} 
  \\
 \end{tabular}
%  \vspace{+2pt}
 \caption{Example results of the proposed AttnGAN. The first row gives the low-to-high resolution images generated by $G_0$, $G1$ and $G_2$ of the AttnGAN; the second and third row shows the top-5 most attended words by $F^{attn}_{1}$ and $F^{attn}_{2}$ of the AttnGAN, respectively. Here, images of $G_0$ and $G_1$ are bilinearly upsampled to have the same size as that of $G_2$ for better visualization.}
 \vspace{-5pt}
 \label{fig:big_examples}
 \end{figure}

To address this issue, we propose an Attentional Generative Adversarial Network (AttnGAN) that allows attention-driven, multi-stage refinement for fine-grained text-to-image generation. The overall architecture of the AttnGAN is illustrated in Figure~\ref{fig:architecture}. The model consists of two novel components.  The first component is an attentional generative network, in which an attention mechanism is developed for the generator to draw different sub-regions of the image by focusing on words that are most relevant to the sub-region being drawn (see Figure~\ref{fig:big_examples}). More specifically, besides encoding the natural language description into a global sentence vector, each word in the sentence is also encoded into a word vector. The generative network utilizes the global sentence vector to generate a low-resolution image in the first stage. In the following stages, it uses the image vector in each sub-region to query word vectors by using an attention layer to form a word-context vector. It then combines the regional image vector and the corresponding word-context vector to form a multimodal context vector, based on which the model generates new image features in the surrounding sub-regions. This effectively yields a higher resolution picture with more details at each stage. The other component in the AttnGAN is a Deep Attentional Multimodal Similarity Model (DAMSM).   With an attention mechanism, the DAMSM is able to compute the similarity between the generated image and the sentence using both the global sentence level information and the fine-grained word level information. Thus, the DAMSM provides an additional fine-grained image-text matching loss for training the generator.

The contribution of our method is threefold. 
(\textit{i}) An Attentional Generative Adversarial Network is proposed for synthesizing images from text descriptions. 
Specifically, two novel components are proposed in the AttnGAN, including the attentional generative network and the DAMSM. 
(\textit{ii}) Comprehensive study is carried out to empirically evaluate the proposed AttnGAN.  Experimental results show that the AttnGAN significantly outperforms previous state-of-the-art GAN models. 
(\textit{iii}) A detailed analysis is performed through visualizing the attention layers of the AttnGAN. 
For the first time, it is demonstrated that the layered conditional GAN is able to automatically attend to relevant words to form the condition for image generation.

%-------------------------------------------------------------------------
\section{Related Work}
\vspace{-5pt}
Generating high resolution images from text descriptions, though very challenging, is important for many practical applications such as art generation and computer-aided design. Recently, great progress has been achieved in this direction with the emergence of deep generative models~\cite{KingmaW14, Oord16, goodfellow2014generative}. Mansimov \etal~\cite{MansimovPBS15} built the alignDRAW model, extending the Deep Recurrent Attention Writer (DRAW)~\cite{Gregor15DRAW} to iteratively draw image patches while attending to the relevant words in the caption. Nguyen \etal~\cite{NguyenYBDC17} proposed an approximate Langevin approach to generate images from captions.  Reed \etal~\cite{Reed17parallel} used conditional PixelCNN~\cite{Oord16} to synthesize images from text with a multi-scale model structure. Compared with other deep generative models, Generative Adversarial Networks (GANs)~\cite{goodfellow2014generative} have shown great performance for generating sharper samples~\cite{Radford15, DentonCSF15, Salimans2016, Christian2016, pix2pix2017}. Reed \etal~\cite{reed2016generative} first showed that the conditional GAN was capable of synthesizing plausible images from text descriptions. Their follow-up work~\cite{reed2016learning} also demonstrated that GAN was able to generate better samples by incorporating additional conditions (\eg, object locations). Zhang \etal~\cite{Han16stackgan, Han17stackgan2} stacked several GANs for text-to-image synthesis and used different GANs to generate images of different sizes. However, all of their GANs are conditioned on the global sentence vector, missing fine-grained word level information for image generation.

The attention mechanism has recently become an integral part of sequence transduction models. It has been successfully used in modeling multi-level dependencies in image captioning~\cite{XuBKCCSZB15}, image question answering~\cite{YangHGDS16} and machine translation~\cite{Dzmitry14}. Vaswani \etal~\cite{Ashish17} also demonstrated that machine translation models could achieve state-of-the-art results by solely using an attention model. In spite of these progress, the attention mechanism has not been explored in GANs for text-to-image synthesis yet. It is worth mentioning that the alignDRAW~\cite{MansimovPBS15} also used LAPGAN~\cite{DentonCSF15} to scale the image to a higher resolution. However, the GAN in their framework was only utilized as a post-processing step without attention.  To our knowledge, the proposed AttnGAN for the first time develops an attention mechanism that enables GANs to generate fine-grained high quality images via  multi-level (\eg, word level and sentence level) conditioning.

%-------------------------------------------------------------------------
\section{Attentional Generative Adversarial Network}
\vspace{-5pt}
\begin{figure*}[tb]
\begin{center}
\includegraphics[width=1.0\linewidth]{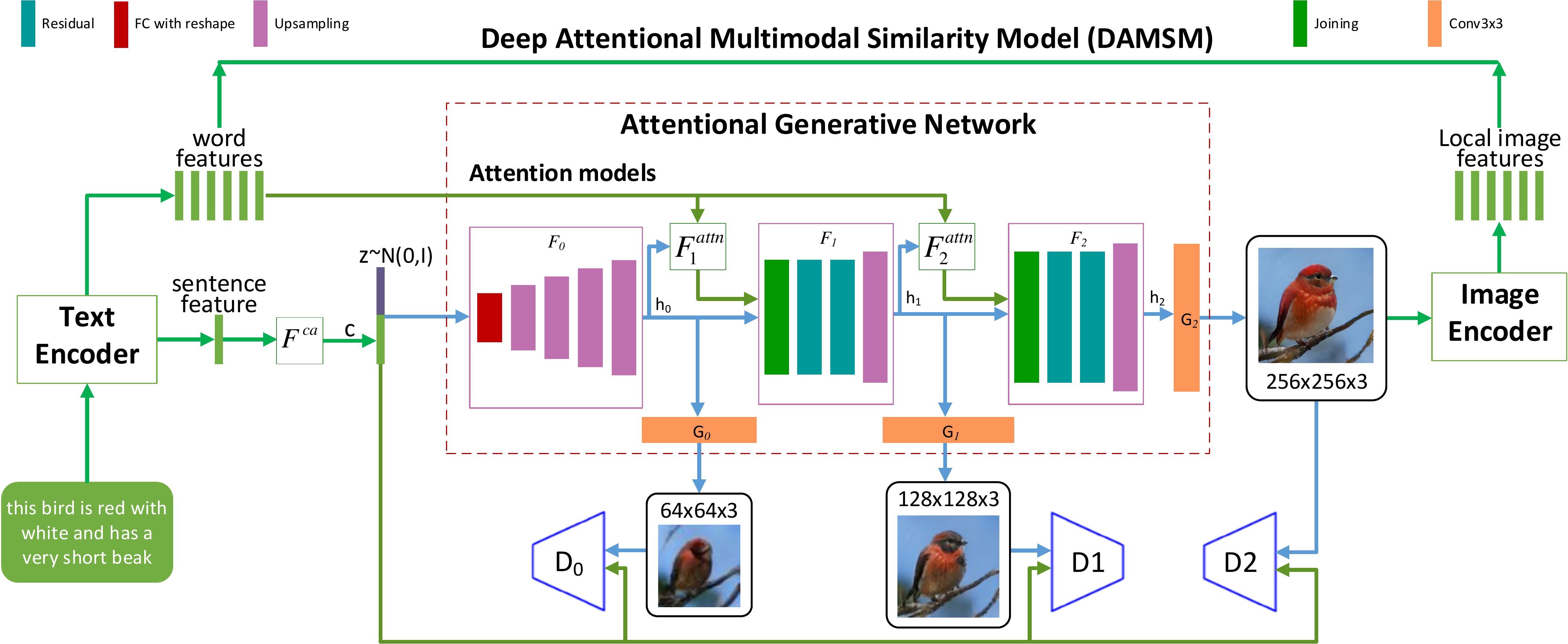}
\end{center}
 \vspace{-10pt}
   \caption{The architecture of the proposed AttnGAN. Each attention model automatically retrieves the conditions (\ie, the most relevant word vectors) for generating different sub-regions of the image; the DAMSM provides the fine-grained image-text matching loss for the generative network.  
   }
%   The details about the basic blocks, such as \textit{Residual} and \textit{Upsampling}, are presented in the supplementary material.
\vspace{-10pt}
	\label{fig:architecture}
\end{figure*}
As shown in Figure~\ref{fig:architecture}, the proposed Attentional Generative Adversarial Network (AttnGAN) has two novel components: the attentional generative network and the deep attentional multimodal similarity model. We will elaborate each of them in the rest of this section.

\subsection{Attentional Generative Network} \label{sec:Fattn}
\vspace{-5pt}
Current GAN-based models for text-to-image generation~\cite{reed2016generative,reed2016learning,Han16stackgan,Han17stackgan2} typically encode the whole-sentence text description into a single vector as the condition for image generation, but lack fine-grained word level information. In this section, we propose a novel attention model that enables the generative network to draw different sub-regions of the image conditioned on words that are most relevant to those sub-regions. 

As shown in Figure~\ref{fig:architecture}, the proposed attentional generative network has $m$ generators ($G_0, G_1, ..., G_{m-1}$), which take the hidden states ($h_0,h_1,...,h_{m-1}$) as input and generate images of small-to-large scales ($\hat{x}_0,\hat{x}_1, ...,\hat{x}_{m-1}$). Specifically,
  \begin{equation}\label{eq:hiddenGs}
   \begin{aligned}
    &h_0 = F_0(z, F^{ca}(\overline{e})); \\ &h_i = F_i(h_{i-1}, F_{i}^{attn}(e, h_{i-1})) \; \text{for} \; i = 1,2, ..., m-1; \; \\ &\hat{x}_i = G_i(h_i).
   \end{aligned}
  \end{equation}
Here, $z$ is a noise vector usually sampled from a standard normal distribution.   $\overline{e}$ is a global sentence vector, and $e$ is the matrix of word vectors.   $F^{ca}$ represents the Conditioning Augmentation~\cite{Han16stackgan} that converts the sentence vector $\overline{e}$ to the conditioning vector.   $F^{attn}_i$ is the proposed attention model at the $i^{th}$ stage of the AttnGAN.   $F^{ca}$, $F^{attn}_i$, $F_i$, and $G_i$ are modeled as neural networks. 

The attention model $F^{attn}(e, h)$ has two inputs: the word features $e \in \mathbb{R}^{D\times T}$ and the image features from the previous hidden layer $h \in \mathbb{R}^{\hat{D}\times N}$. The word features are first converted into the common semantic space of the image features by adding a new perceptron layer, \ie, $e^{\prime} = Ue$, where $U \in \mathbb{R}^{\hat{D}\times D}$. Then, a word-context vector is computed for each sub-region of the image based on its hidden features $h$ (query). Each column of $h$ is a feature vector of a sub-region of the image. For the $j^{th}$ sub-region, its word-context vector is a dynamic representation of word vectors relevant to $h_j$, which is calculated by
 \begin{equation}\label{eq:AttnGAN}
  c_j = \sum_{i=0}^{T-1}\beta_{j,i}e^{\prime}_{i}, \; \; \text{where} \; \beta_{j,i} = \frac{\exp(s_{j,i}^{\prime})}{\sum_{k=0}^{T-1}{\exp(s_{j,k}^{\prime})}}, 
 \end{equation}
$s_{j,i}^{\prime}= h_j^T e^{\prime}_i$, and $\beta_{j,i}$ indicates the weight the model attends to the $i^{th}$ word when generating the $j^{th}$  sub-region of the image.  We then donate the word-context matrix for image feature set $h$ by $F^{attn}(e, h) = (c_0, c_1,..., c_{N-1}) \in \mathbb{R}^{\hat{D}\times N}$. Finally, image features and the corresponding word-context features are combined to generate images at the next stage.

To generate realistic images with multiple levels (\ie, sentence level and word level) of conditions, the final objective function of the attentional generative network is defined as
 \begin{equation}\label{eq:final-LG}
  \mathcal{L} = \mathcal{L}_{G} + \lambda\mathcal{L}_{DAMSM}, \; \; \text{where} \; \mathcal{L}_{G} = \sum_{i=0}^{m-1} \mathcal{L}_{G_{i}}.
 \end{equation}
Here, $\lambda$ is a hyperparameter to balance the two terms of Eq.~(\ref{eq:final-LG}). The first term is the GAN loss that jointly approximates conditional and unconditional distributions~\cite{Han17stackgan2}. At the $i^{th}$ stage of the AttnGAN, the generator $G_i$ has a corresponding discriminator $D_i$. The adversarial loss for $G_i$ is defined as
 \begin{equation}\label{eq:hybrid-LGi}
  \footnotesize
   \begin{aligned}
    \mathcal{L}_{G_{i}} &= \underbrace{-\frac{1}{2} \mathbb{E}_{\hat{x}_{i} \sim {p_{G_{i}}}} [\log(D_{i}(\hat{x}_{i})] }_\text{unconditional loss} \; \underbrace{-\frac{1}{2} \mathbb{E}_{\hat{x}_{i} \sim {p_{G_{i}}}} [\log(D_{i}(\hat{x}_{i}, \overline{e})]}_\text{conditional loss},
   \end{aligned}
 \end{equation}
where the unconditional loss determines whether the image is real or fake while the conditional loss determines whether the image and the sentence match or not. 

Alternately to the training of $G_i$, each discriminator $D_i$ is trained to classify the input into the class of real or fake by minimizing the cross-entropy loss defined by 
\begin{equation}\label{eq:hybrid-LDi}
\scriptsize
\begin{aligned}
\mathcal{L}_{D_{i}} &= \underbrace{-\frac{1}{2}  \mathbb{E}_{x_{i} \sim {p_{data_i}}} [\log D_{i}(x_{i})] \; -\frac{1}{2} \mathbb{E}_{\hat{x}_{i} \sim {p_{G_{i}}}} [\log(1 - D_{i}(\hat{x}_{i})] }_\text{unconditional loss} + \\   
&\;\;\;\; \underbrace{-\frac{1}{2}  \mathbb{E}_{x_{i} \sim {p_{data_i}}} [\log D_{i}(x_{i}, \overline{e})] \; -\frac{1}{2} \mathbb{E}_{\hat{x}_{i} \sim {p_{G_{i}}}} [\log(1 - D_{i}(\hat{x}_{i}, \overline{e})] }_\text{conditional loss},
\end{aligned}
\end{equation}
where $x_i$ is from the true image distribution $p_{data_i}$ at the $i^{th}$ scale, and $\hat{x}_i$ is from the model distribution $p_{G_{i}}$ at the same scale. Discriminators of the AttnGAN are structurally disjoint, so they can be trained in parallel and each of them focuses on a single image scale. 

The second term of Eq.~(\ref{eq:final-LG}), $\mathcal{L}_{DAMSM}$, is a word level fine-grained image-text matching loss computed by the DAMSM, which will be elaborated in Subsection~\ref{sec:DAMSM}.

\subsection{Deep Attentional Multimodal Similarity Model} \label{sec:DAMSM}
\vspace{-5pt}
The DAMSM learns two neural networks that map sub-regions of the image and words of the sentence to a common semantic space, thus measures the image-text similarity at the word level to compute a fine-grained loss for image generation.

\textbf{The text encoder }{
is a bi-directional Long Short-Term Memory (LSTM)~\cite{SchusterP97} that extracts semantic vectors from the text description. In the bi-directional LSTM, each word corresponds to two hidden states, one for each direction. Thus, we concatenate its two hidden states to represent the semantic meaning of a word. The feature matrix of all words is indicated by $e \in \mathbb{R}^{D\times T}$. Its $i^{th}$ column $e_i$ is the feature vector for the $i^{th}$ word. $D$ is the dimension of the word vector and $T$ is the number of words. Meanwhile, the last hidden states of the bi-directional LSTM are concatenated to be the global sentence vector, denoted by $\overline{e} \in \mathbb{R}^{D}$. 
}

\textbf{The image encoder }{
is a Convolutional Neural Network (CNN) that maps images to semantic vectors. The intermediate layers of the CNN learn local features of different sub-regions of the image, while the later layers learn global features of the image. More specifically, our image encoder is built upon the Inception-v3 model~\cite{SzegedyVISW16} pretrained on ImageNet~\cite{ILSVRC15}. We first rescale the input image to be 299$\times$299 pixels. And then, we extract the local feature matrix $f \in \mathbb{R}^{768\times289}$ (reshaped from 768$\times$17$\times$17) from the ``$mixed\_6e$'' layer of Inception-v3. Each column of $f$ is the feature vector of a sub-region of the image. 768 is the dimension of the local feature vector, and 289 is the number of sub-regions in the image. Meanwhile, the global feature vector $\overline{f} \in \mathbb{R}^{2048}$ is extracted from the last average pooling layer of Inception-v3. Finally, we convert the image features to a common semantic space of text features by adding a perceptron layer: 
 \begin{equation}\label{eq:localFtr}
  v = W f\,, \, \, \, \,\, \, \, \, \overline{v} = \overline{W}\,\overline{f},
 \end{equation}
where $v \in \mathbb{R}^{D\times289}$ and its $i^{th}$ column $v_{i}$ is the visual feature vector for the $i^{th}$ sub-region of the image; and $\overline{v} \in \mathbb{R}^{D}$ is the global vector for the whole image. $D$ is the dimension of the multimodal (\ie, image and text modalities) feature space. For efficiency, all parameters in layers built from the Inception-v3 model are fixed, and the parameters in newly added layers are jointly learned with the rest of the network. 
}

\textbf{The attention-driven image-text matching score }{
is designed to measure the matching of an image-sentence pair based on an attention model between the image and the text. 

We first calculate the similarity matrix for all possible pairs of words in the sentence and sub-regions in the image by
\begin{equation}\label{eq:similarity}
s = e^T\,v,
\end{equation}
where $s \in \mathbb{R}^{T\times 289}$ and $s_{i,j}$ is the dot-product similarity between the $i^{th}$ word of the sentence and the $j^{th}$ sub-region of the image. We find that it is beneficial to normalize the similarity matrix as follows
\begin{equation}\label{eq:similarity_normalized}
  \overline{s}_{i,j} = \frac{\exp(s_{i,j})}{\sum_{k=0}^{T-1}{\exp(s_{k,j})}}. 
\end{equation}

Then, we build an attention model to compute a region-context vector for each word (query). The region-context vector $c_{i}$ is a dynamic representation of the image's sub-regions related to the $i^{th}$ word of the sentence. It is computed as the weighted sum over all regional visual vectors, \ie,
\begin{equation}\label{eq:attention}
  c_i = \sum_{j=0}^{288}\alpha_{j}v_{j}, \; \; \text{where} \; \alpha_{j} = \frac{\exp(\gamma_{1} \overline{s}_{i,j})}{\sum_{k=0}^{288}{\exp(\gamma_{1} \overline{s}_{i,k}})}.
\end{equation}
Here, $\gamma_{1}$ is a factor that determines how much attention is paid to features of its relevant sub-regions when computing the region-context vector for a word.

Finally, we define the relevance between the $i^{th}$ word and the image using the cosine similarity between $c_i$ and $e_i$, \ie, $R(c_i, e_i) = (c_i^T e_i) / (||c_i|| ||e_i||)$. 
Inspired by the minimum classification error formulation in speech recognition (see, \eg, \cite{Juang97minimum, he08discriminatice}), the \textit{attention-driven image-text matching score} between the entire image ($Q$) and the whole text description ($D$) is defined as
 \begin{equation}\label{eq:relevance}
  R(Q, D) = \log \Big(\sum_{i=1}^{T-1} \exp(\gamma_{2} R(c_i, e_i)) \Big)^{\frac{1}{\gamma_{2}}}, 
 \end{equation}
where $\gamma_{2}$ is a factor that determines how much to magnify the importance of the most relevant word-to-region-context pair. When $\gamma_{2}\to\infty$, $R(Q, D)$ approximates to $\max_{i=1}^{T-1} R(c_i, e_i)$.
}

\textbf{The DAMSM loss} {
is designed to learn the attention model in a semi-supervised manner, in which the only supervision is the matching between entire images and whole sentences (a sequence of words). Similar to \cite{FangGISDDGHMPZZ15,Huang13LDS}, for a batch of image-sentence pairs $\{(Q_i, D_i)\}_{i=1}^M$, the posterior probability of sentence $D_i$ being matching with image $Q_i$ is computed as
 \begin{equation}\label{eq:relevant_prob}
  P(D_i|Q_i) = \frac{\exp(\gamma_{3} R(Q_i, D_i))}{\sum_{j=1}^M{\exp(\gamma_{3} R(Q_i, D_{j}))}},
 \end{equation}
where $\gamma_{3}$ is a smoothing factor determined by experiments. In this batch of sentences, only $D_i$ matches the image $Q_i$, and treat all other $M-1$ sentences as mismatching descriptions. Following \cite{FangGISDDGHMPZZ15,Huang13LDS}, we define the loss function as the negative log posterior probability that the images are matched with their corresponding text descriptions (ground truth), \ie, 
\begin{equation}\label{eq:L1}
\mathcal{L}^{w}_{1} = -\sum_{i=1}^M \log P(D_i|Q_i),
\end{equation}
where `w' stands for ``word''.

Symmetrically, we also minimize 
\begin{equation}\label{eq:L2}
\mathcal{L}^{w}_{2} = -\sum_{i=1}^M \log P(Q_i|D_i),
\end{equation}
where $P(Q_i|D_i) = \frac{\exp(\gamma_{3} R(Q_i, D_i))}{\sum_{j=1}^M {\exp(\gamma_{3} R(Q_{j}, D_i))}}$ is the posterior probability that sentence $D_i$ is matched with its corresponding image $Q_i$. 
If we redefine Eq.~\eqref{eq:relevance} by $R(Q, D) = \big(\overline{v}^T \overline{e}\big) / \big(||\overline{v}|| ||\overline{e}||\big)$ and substitute it to Eq.~\eqref{eq:relevant_prob}, \eqref{eq:L1} and \eqref{eq:L2}, we can obtain loss functions $\mathcal{L}^{s}_{1}$ and $\mathcal{L}^{s}_{2}$ (where `s' stands for ``sentence'') using the sentence vector $\overline{e}$ and the global image vector $\overline{v}$.

Finally, the DAMSM loss is defined as 
 \begin{equation}\label{eq:L_DAMSM}
  % \begin{aligned}
   \mathcal{L}_{DAMSM} = \mathcal{L}^{w}_{1} + \mathcal{L}^{w}_{2} + \mathcal{L}^{s}_{1} + \mathcal{L}^{s}_{2}.
  % \end{aligned}
 \end{equation}
Based on experiments on a held-out validation set, we set the hyperparameters in this section as: $\gamma_{1} = 5$, $\gamma_{2} = 5$, $\gamma_{3} = 10$ and $M=50$. Our DAMSM is pretrained~\footnote{We also finetuned the DAMSM with the whole network, however the performance was not improved.} by minimizing $\mathcal{L}_{DAMSM}$ using real image-text pairs. Since the size of images for pretraining DAMSM is not limited by the size of images that can be generated, real images of size 299$\times$299 are utilized. In addition, the pretrained text-encoder in the DAMSM provides visually-discriminative word vectors learned from image-text paired data for the attentional generative network. In comparison, conventional word vectors pretrained on pure text data are often not visually-discriminative, \eg, word vectors of different colors, such as red, blue, yellow, etc., are often clustered together in the vector space, due to the lack of grounding them to the actual visual signals.
}

In sum, we propose two novel attention models, the attentional generative network and the DAMSM, which play different roles in the AttnGAN. (\textit{i}) The attention mechanism in the generative network (see Eq.~\ref{eq:AttnGAN}) enables the AttnGAN to automatically select word level condition for generating different sub-regions of the image. (\textit{ii}) With an attention mechanism (see Eq.~\ref{eq:attention}), the DAMSM is able to compute the fine-grained text-image matching loss $\mathcal{L}_{DAMSM}$. It is worth mentioning that, $\mathcal{L}_{DAMSM}$ is applied only on the output of the last generator $G_{m-1}$, because the eventual goal of the AttnGAN is to generate large images by the last generator. We tried to apply $\mathcal{L}_{DAMSM}$ on images of all resolutions generated by ($G_0, G_1, ..., G_{m-1}$).  However, the performance was not improved but the computational cost was increased.

\section{Experiments}
\vspace{-5pt}
{
Extensive experimentation is carried out to evaluate the proposed AttnGAN. 
We first study the important components of the AttnGAN, including the attentional generative network and the DAMSM. 
Then, we compare our AttnGAN with previous state-of-the-art GAN models for text-to-image synthesis~\cite{Han16stackgan,Han17stackgan2,reed2016generative,reed2016learning,NguyenYBDC17}. 
}

\textbf{Datasets. }{
Same as previous text-to-image methods~\cite{Han16stackgan,Han17stackgan2,reed2016generative,reed2016learning}, our method is evaluated on CUB~\cite{WahCUB_200_2011} and COCO~\cite{LinMBHPRDZ14} datasets. 
We preprocess the CUB dataset according to the method in \cite{Han16stackgan}.  
Table~\ref{tab:dataset} lists the statistics of datasets. 
}

 \begin{table}[bt]
  \begin{center}
  \small
   \begin{tabular}{|l|c|c|c|c|}
    \hline
     \multirow{2}{4em}{Dataset}&\multicolumn{2}{c|}{CUB~\cite{WahCUB_200_2011}} &\multicolumn{2}{c|}{COCO~\cite{LinMBHPRDZ14}} \\
    \cline{2-5}
     &train &test &train &test \\
    \hline
     \#samples&8,855 &2,933 &80k &40k \\
    \hline
     caption/image&10 &10 &5 &5 \\
    \hline
   \end{tabular}
  \end{center}
  \vspace{-5pt}
  \caption{Statistics of datasets.}
  \vspace{-10pt}
  \label{tab:dataset} 
 \end{table}

\textbf{Evaluation. }{
Following Zhang \etal~\cite{Han16stackgan}, we use the inception score \cite{Salimans2016} as the quantitative evaluation measure. Since the inception score cannot reflect whether the generated image is well conditioned on the given text description, we propose to use R-precision, a common evaluation metric for ranking retrieval results, as a complementary evaluation metric for the text-to-image synthesis task. If there are $R$ relevant documents for a query, we examine the top $R$ ranked retrieval results of a system, and find that $r$ are relevant, and then by definition, the R-precision is ${r}/{R}$. More specifically, we conduct a retrieval experiment, \ie, we use generated images to query their corresponding text descriptions. First, the image and text encoders learned in our pretrained DAMSM are utilized to extract global feature vectors of the generated images and the given text descriptions. And then, we compute cosine similarities between the global image vectors and the global text vectors. Finally, we rank candidate text descriptions for each image in descending similarity and find the top $r$ relevant descriptions for computing the R-precision. To compute the inception score and the R-precision, each model generates 30,000 images from randomly selected unseen text descriptions. The candidate text descriptions for each query image consist of one ground truth (\ie, $R=1$) and 99 randomly selected mismatching descriptions.

Besides quantitative evaluation, we also qualitatively examine the samples generated by our models.  Specifically, we visualize the intermediate results with attention learned by the attention models $F^{attn}$. As defined in Eq.~(\ref{eq:AttnGAN}), weights $\beta_{j,i}$ indicates which words the model attends to when generating a sub-region of the image, and $\sum_{i=0}^{T-1} \beta_{j,i} = 1$. 
We suppress the less-relevant words for an image's sub-region via
 \begin{equation}\label{eq:suppress}
    \hat{\beta}_{j,i} = 
        \begin{cases}
            \beta_{j,i}, & \text{if } \beta_{j,i} > 1/T,\\
                0,              & \text{otherwise.}
        \end{cases}
 \end{equation}
For better visualization, we fix the word and compute its attention weights with $N$ different sub-regions of an image, $\hat{\beta}_{0,i}, \hat{\beta}_{1,i}, ..., \hat{\beta}_{N-1,i}$. We reshape the $N$ attention weights to $\sqrt{N}\times\sqrt{N}$ pixels, which are then upsampled with Gaussian filters to have the same size as the generated images. Limited by the length of the paper, we only visualize the top-5 most attended words (\ie., words with top-5 highest $\sum_{j=0}^{N-1} \hat{\beta}_{j,i}$ values) for each attention model. 
}

\begin{figure}[bt]
\begin{center}
	\includegraphics[width=0.48\linewidth]{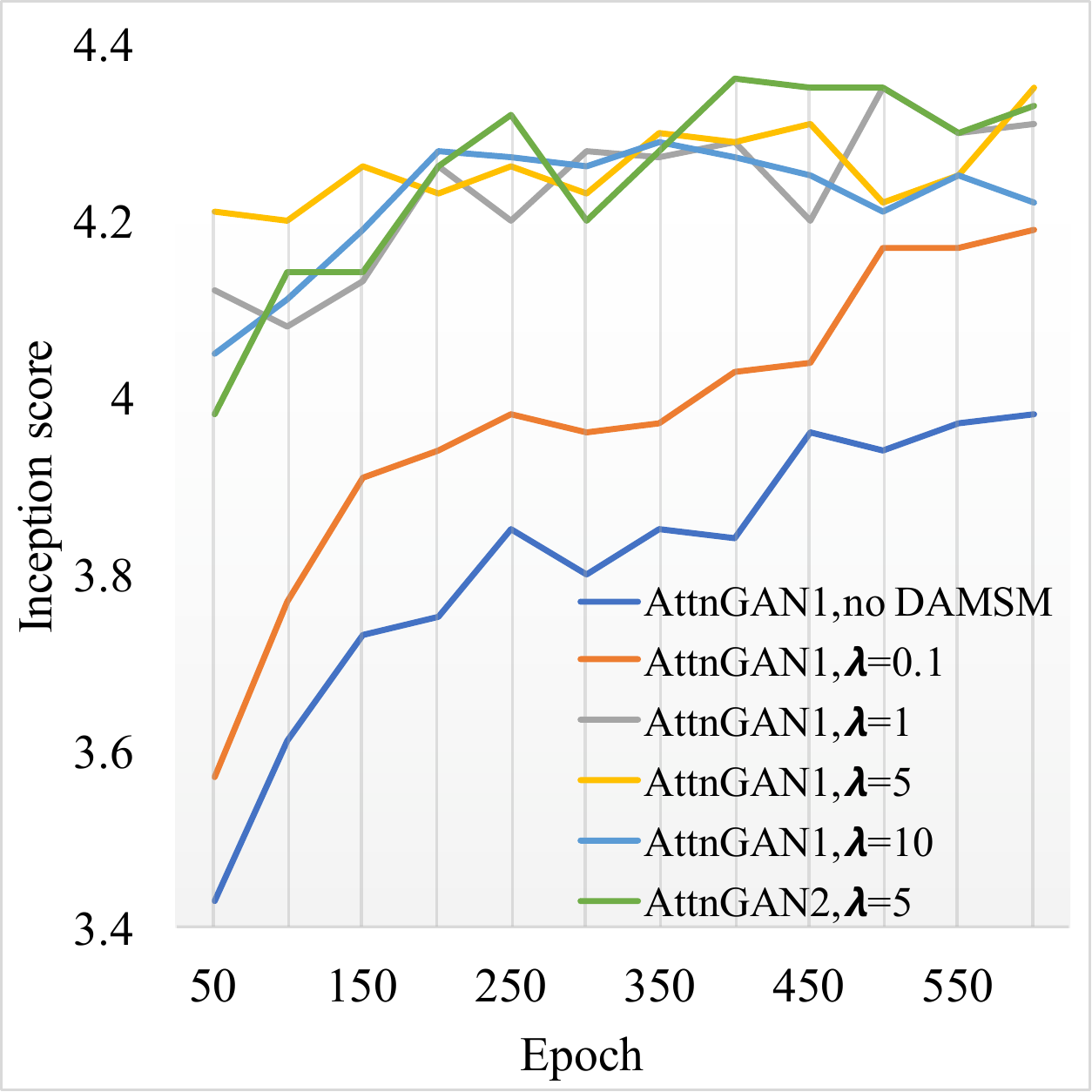}
	\includegraphics[width=0.48\linewidth]{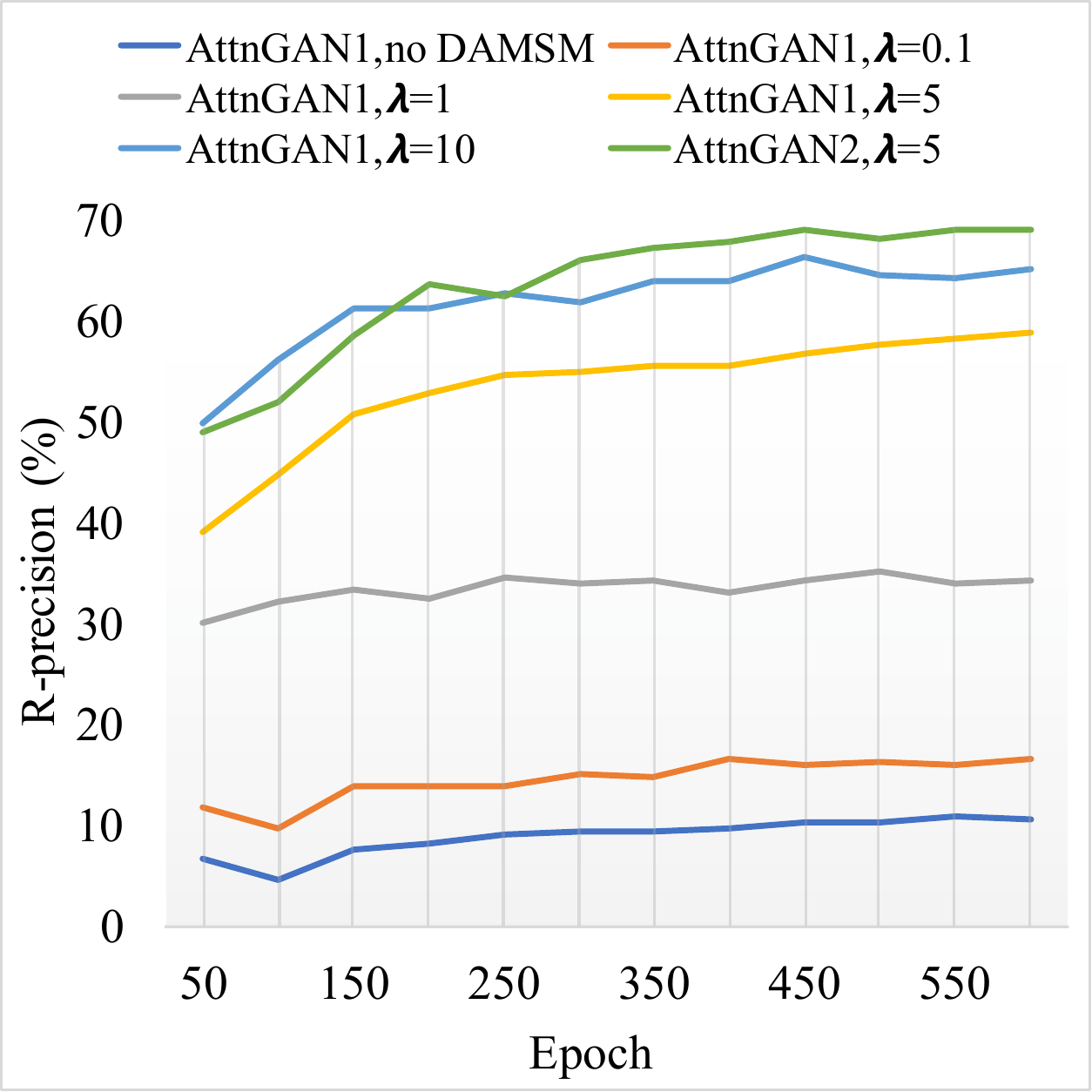} \\ \vspace{+5pt}
	\includegraphics[width=0.48\linewidth]{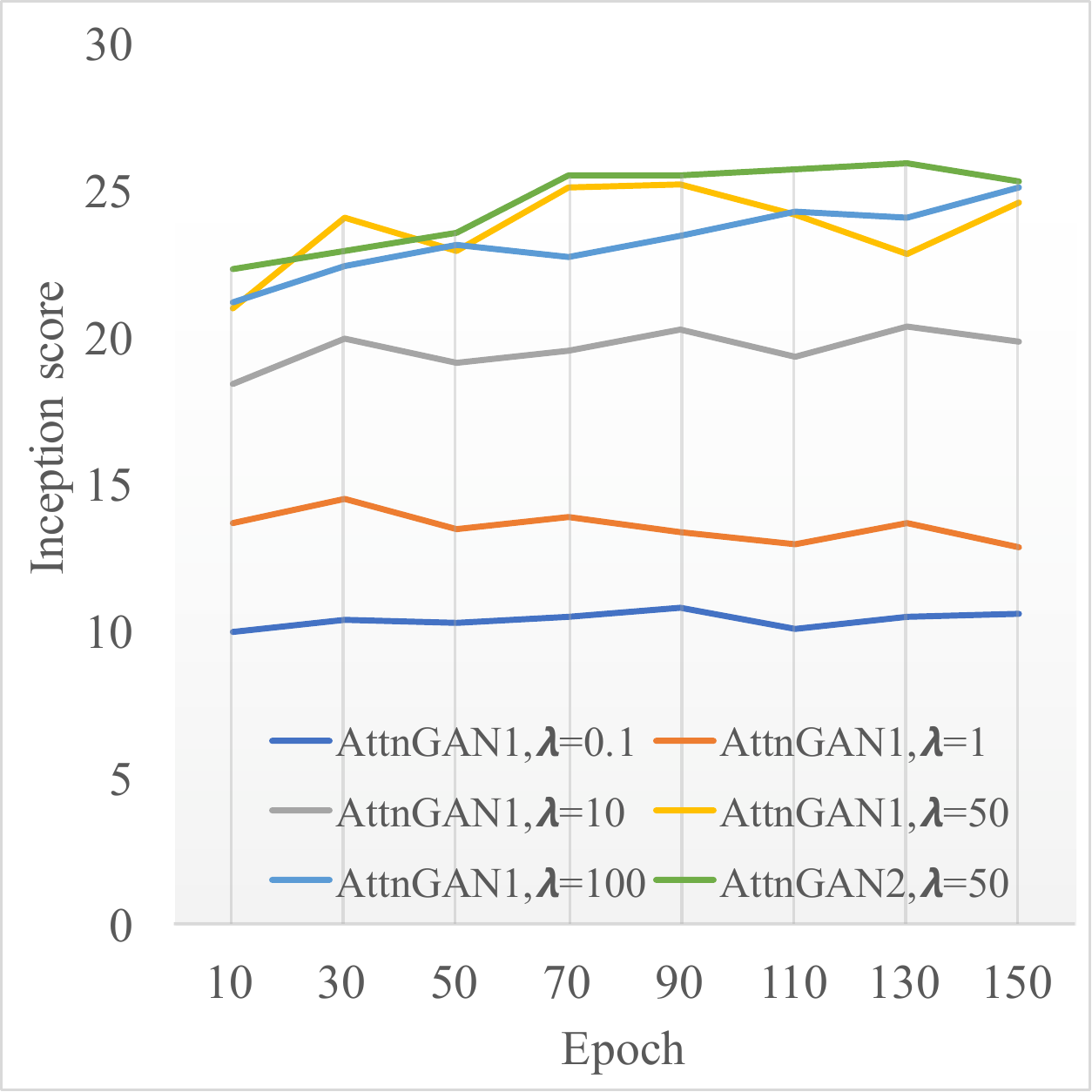}
	\includegraphics[width=0.48\linewidth]{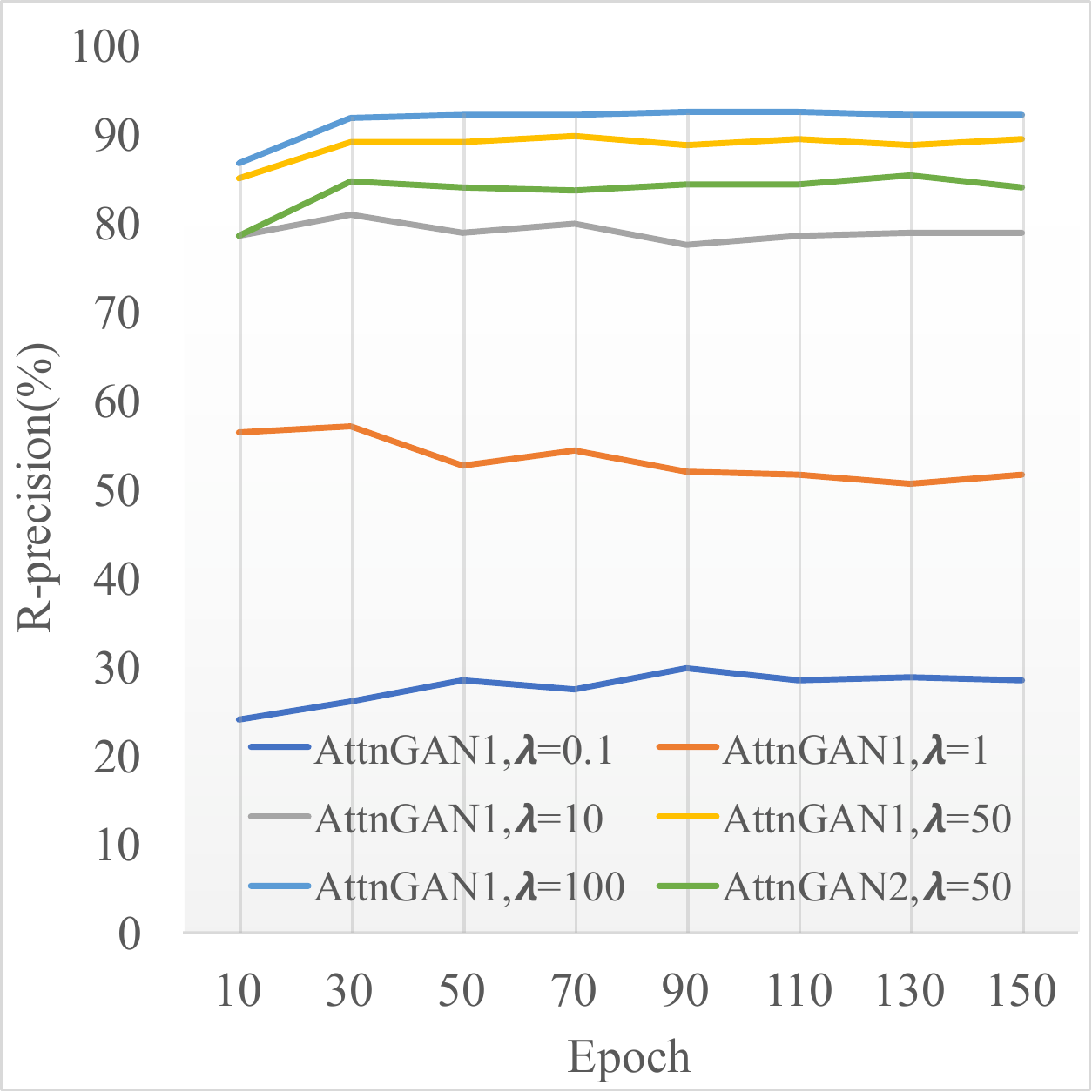}
\end{center}
\vspace{-8pt}
   \caption{Inception scores and R-precision rates by our AttnGAN and its variants at different epochs on CUB (top) and COCO (bottom) test sets. For the text-to-image synthesis task, $R=1$.}
% \vspace{-10pt}
\label{fig:component}
\end{figure}
 \begin{table}[bt]
  \begin{center}
    \small
    % \scriptsize
    % \footnotesize
   \begin{tabular}{l|l|l}
    % \hline
     Method & inception score & R-precision(\%)\\
    \hline
     AttnGAN1,~no DAMSM       &3.98~$\pm$~.04  & 10.37$\pm$~5.88 \\
    % \hline
     AttnGAN1,~$\lambda=0.1$  &4.19~$\pm$~.06  & 16.55$\pm$~4.83\\
     AttnGAN1,~$\lambda=1$    &4.35~$\pm$~.05  & 34.96$\pm$~4.02  \\
     AttnGAN1,~$\lambda=5$    &4.35~$\pm$~.04  & 58.65$\pm$~5.41  \\
     AttnGAN1,~$\lambda=10$   &4.29~$\pm$~.05  & 63.87$\pm$~4.85  \\
    \hline
     \textbf{AttnGAN2,~$\lambda=5$} &\textbf{4.36~$\pm$~.03}   & \textbf{67.82~$\pm$~4.43}  \\
     \hline
      \textbf{AttnGAN2,~$\lambda=50$} &\multirow{2}{4em}{\textbf{25.89~$\pm$~.47}}   & \multirow{2}{4em}{\textbf{85.47~$\pm$~3.69}} \\
      ~~~~~~~~~\textbf{(COCO)} & & \\  
    \hline
   \end{tabular}
  \end{center}
   % \vspace{-5pt}
   \caption{The best inception score and the corresponding R-precision rate of each AttnGAN model on CUB (top six rows) and COCO (the last row) test sets. More results in Figure~\ref{fig:component}.}
    \vspace{-10pt}
  \label{tab:component} 
 \end{table}
\begin{figure*}[h]
 \small
 \centering
 \begin{tabular}{|@{\hspace{1mm}}p{0.095\textwidth}@{\hspace{1mm}}p{0.095\textwidth}@{\hspace{1mm}}p{0.095\textwidth}@{\hspace{1mm}}p{0.095\textwidth}@{\hspace{1mm}}p{0.095\textwidth}|p{0.095\textwidth}@{\hspace{1mm}}p{0.095\textwidth}@{\hspace{1mm}}p{0.095\textwidth}@{\hspace{1mm}}p{0.095\textwidth}@{\hspace{1mm}}p{0.095\textwidth}@{\hspace{1mm}}|}
 \hline
  \multicolumn{5}{|@{\hspace{1mm}}c|}{\shortstack{\\{\color{red}the bird} has a {\color{red} yellow} crown and a {\color{blue} black} eyering that {\color{red} is round}}} &
  \multicolumn{5}{c@{\hspace{1mm}}|}{\shortstack{\\ {\color{red} this bird has} a  {\color{blue} green} crown {\color{blue} black} primaries and a  {\color{red} white belly}}} 
  \\
  \multicolumn{5}{|@{\hspace{1mm}}c|}{\includegraphics[width=0.159\linewidth]{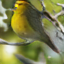} \includegraphics[width=0.159\linewidth]{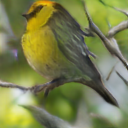} \includegraphics[width=0.159\linewidth]{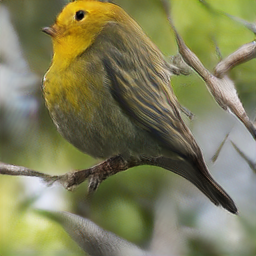}} & \multicolumn{5}{c@{\hspace{1mm}}|}{\includegraphics[width=0.159\linewidth]{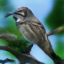} \includegraphics[width=0.159\linewidth]{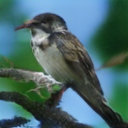} \includegraphics[width=0.159\linewidth]{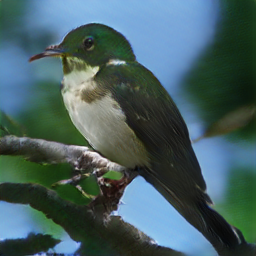}}
  \\
  \shortstack{\color{red}~1:bird} &\shortstack{\color{red}4:yellow} &\shortstack{\color{red}0:the} &\shortstack{\color{red}12:round} &\shortstack{\color{red}11:is} 
  &\shortstack{\color{red}~1:bird} &\shortstack{\color{red}0:this} &\shortstack{\color{red}2:has} &\shortstack{\color{red}11:belly} &\shortstack{\color{red}10:white}
  \\
  \multicolumn{5}{|@{\hspace{1mm}}c|}{\includegraphics[width=0.489\linewidth]{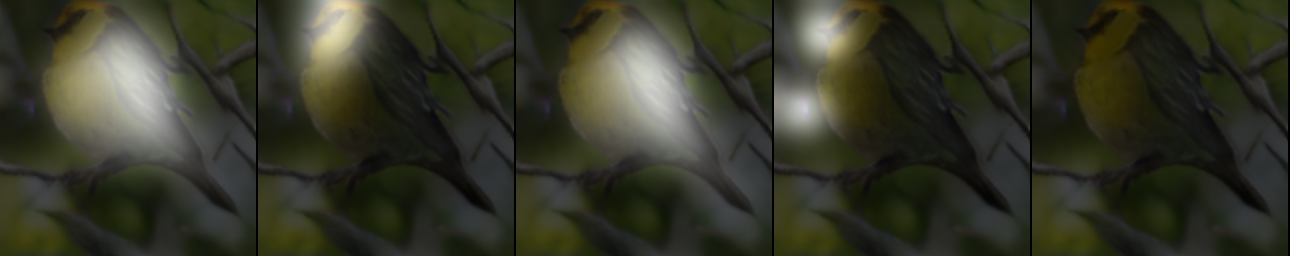}} & \multicolumn{5}{c@{\hspace{1mm}}|}{\includegraphics[width=0.489\linewidth]{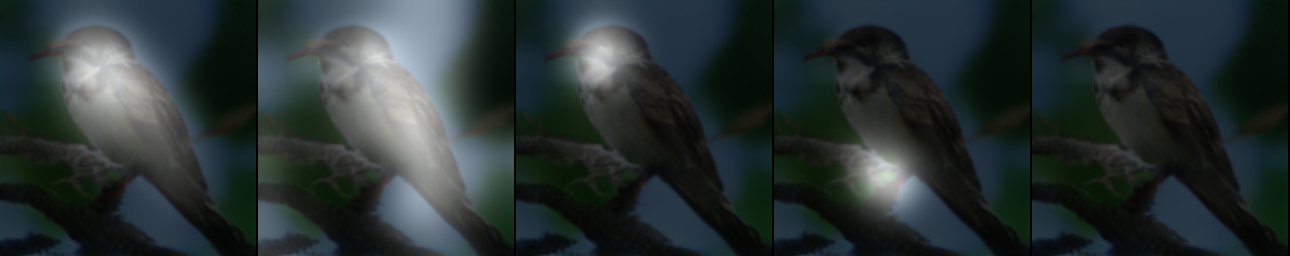}} 
%   \vspace{-10pt}
  \\
  \shortstack{\color{blue}~1:bird} &\shortstack{\color{blue}4:yellow} &\shortstack{\color{blue}0:the} &\shortstack{\color{blue}8:black} &\shortstack{\color{blue}12:round} 
  &\shortstack{\color{blue}~6:black} &\shortstack{\color{blue}4:green} &\shortstack{\color{blue}10:white} &\shortstack{\color{blue}0:this} &\shortstack{\color{blue}1:bird}
  \\
  \multicolumn{5}{|@{\hspace{1mm}}c|}{\includegraphics[width=0.489\linewidth]{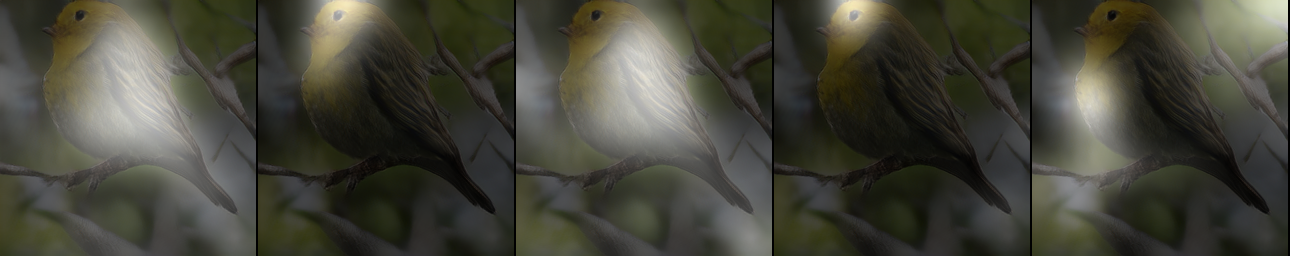}} & \multicolumn{5}{c@{\hspace{1mm}}|}{\includegraphics[width=0.489\linewidth]{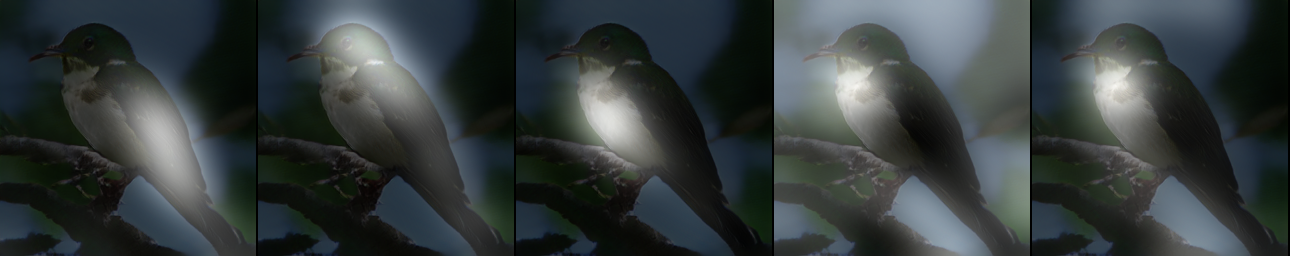}} 
  \\
  \hline
  % ###################################
  \hline
  \multicolumn{5}{|@{\hspace{1mm}}c|}{\shortstack{\\{\color{red}a} photo of a homemade {\color{red}swirly} {\color{blue}pasta} {\color{red}with broccoli} {\color{blue}carrots} {\color{red}and} onions}} &
   \multicolumn{5}{c@{\hspace{1mm}}|}{\shortstack{\\{\color{red}a fruit} stand display with {\color{red}bananas and kiwi}}}
  \\
  \multicolumn{5}{|@{\hspace{1mm}}c|}{\includegraphics[width=0.159\linewidth]{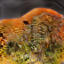} \includegraphics[width=0.159\linewidth]{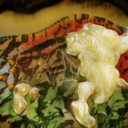} \includegraphics[width=0.159\linewidth]{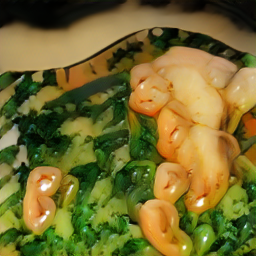}}&
  \multicolumn{5}{c@{\hspace{1mm}}|}{\includegraphics[width=0.159\linewidth]{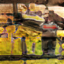} \includegraphics[width=0.159\linewidth]{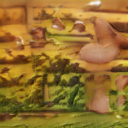} \includegraphics[width=0.159\linewidth]{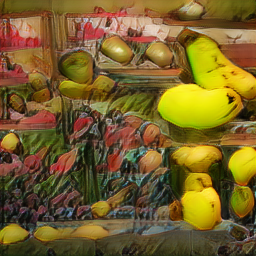}}
  \\
  \shortstack{\color{red}~0:a} &\shortstack{\color{red}7:with} &\shortstack{\color{red}5:swirly} &\shortstack{\color{red}8:broccoli}
  &\shortstack{\color{red}10:and}
  &\shortstack{\color{red}~0:a} &\shortstack{\color{red}6:and} &\shortstack{\color{red}1:fruit} &\shortstack{\color{red}7:kiwi} &\shortstack{\color{red}5:bananas}
  \\
  \multicolumn{5}{|@{\hspace{1mm}}c|}{\includegraphics[width=0.489\linewidth]{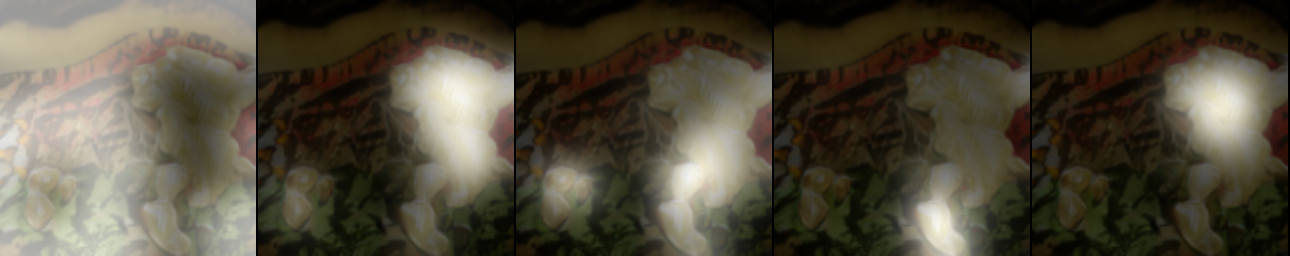}} &
  \multicolumn{5}{c@{\hspace{1mm}}|}{\includegraphics[width=0.489\linewidth]{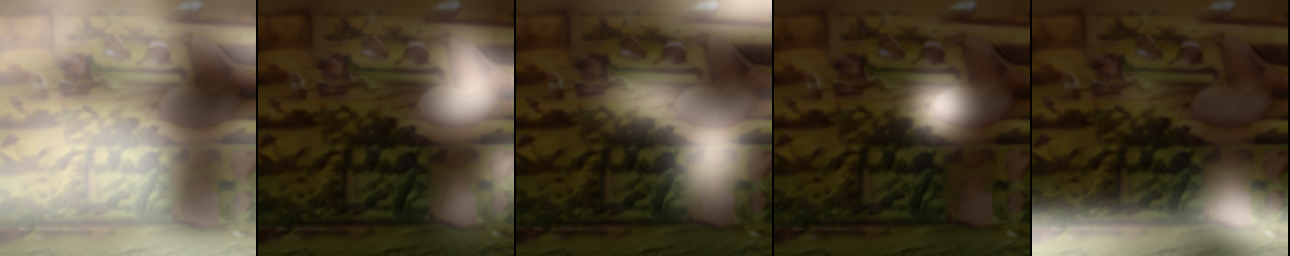}}
  \\
  \shortstack{\color{blue}~8:broccoli} &\shortstack{\color{blue}6:pasta} &\shortstack{\color{blue}0:a} &\shortstack{\color{blue}9:carrot} &\shortstack{\color{blue}5:swirly}
  &\shortstack{\color{blue}~0:a} &\shortstack{\color{blue}5:bananas} &\shortstack{\color{blue}1:fruit} &\shortstack{\color{blue}7:kiwi} &\shortstack{\color{blue}6:and}
  \\
  \multicolumn{5}{|@{\hspace{1mm}}c|}{\includegraphics[width=0.489\linewidth]{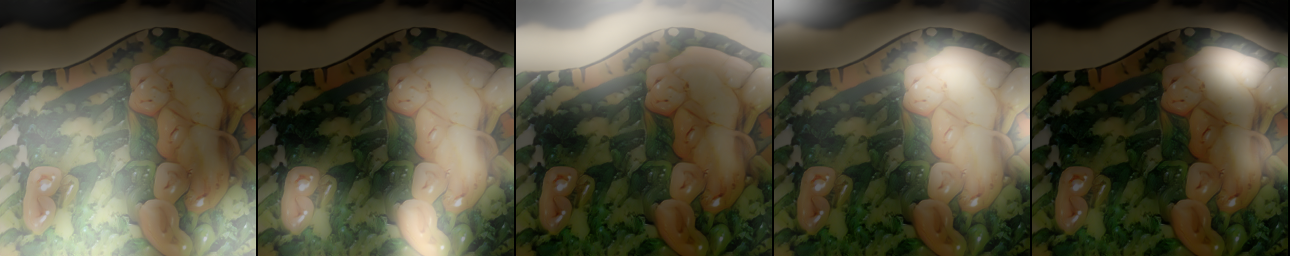}} &
  \multicolumn{5}{c@{\hspace{1mm}}|}{\includegraphics[width=0.489\linewidth]{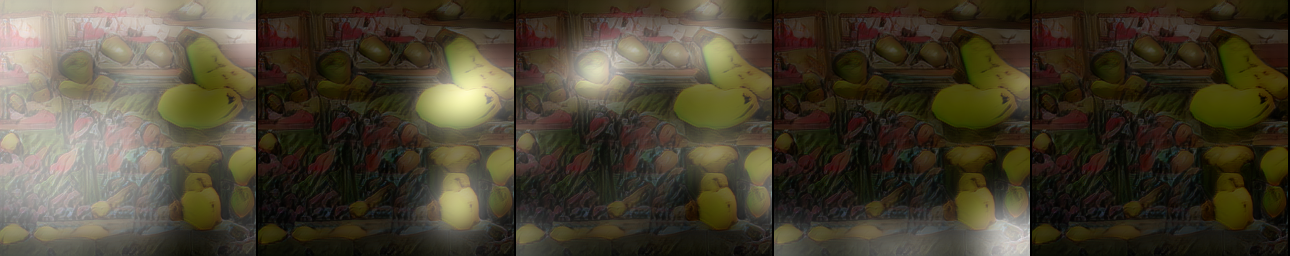}} 
  \\
  \hline
%  COCO1_epoch80: \shortstack{\\{\color{red} the} {\color{blue}girl} {\color{red}is surfing} a small wave in {\color{red}the water}}
% COCO1: \shortstack{\\{\color{red}a} {\color{blue}herd} of {\color{blue}sheep} grazing on \\{\color{red}a lush green filed}}
%  COCO2: \shortstack{\\{\color{red}an} image of a {\color{red}girl} eating \\{\color{red}a} large {\color{blue}slice} {\color{red}of pizza}}
%   COCO4: \shortstack{\\{\color{red}a} {\color{blue}fluffy} black {\color{red}cat sitting on} \\{\color{blue}a laptop} computer {\color{red}keyboard}}
 \end{tabular}
 \vspace{+3pt}
 \caption{Intermediate results of our AttnGAN on CUB (top) and COCO (bottom) test sets. In each block, the first row gives 64$\times$64 images by $G_0$, 128$\times$128 images by $G1$ and 256$\times$256 images by $G_2$ of the AttnGAN; the second and third row shows the top-5 most attended words by $F^{attn}_{1}$ and $F^{attn}_{2}$ of the AttnGAN, respectively. Refer to the supplementary material for more examples.}
 % The top-5 words attended by $F^{attn}_{1}$ is marked in red while the newly attended words by $F^{attn}_{2}$ are marked in blue.
%  \vspace{+10pt}
 \label{fig:examples}
 \end{figure*}
\begin{figure}[h]
 \small
 \centering
 \begin{tabular}{@{\hspace{1mm}}c@{\hspace{1mm}}c@{\hspace{1mm}}c@{\hspace{1mm}}c@{\hspace{1mm}}c@{\hspace{1mm}}c@{\hspace{1mm}}c@{\hspace{1mm}}}
%  \hline
  \multicolumn{7}{c}{this bird has wings that are {\color{red}black} and has a {\color{red}white} belly}
  \\
  \includegraphics[width=0.13\linewidth]{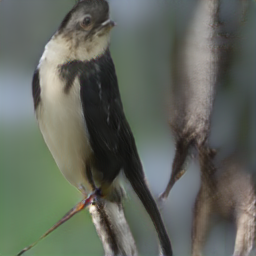}&
  \includegraphics[width=0.13\linewidth]{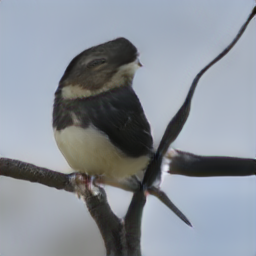}&
  \includegraphics[width=0.13\linewidth]{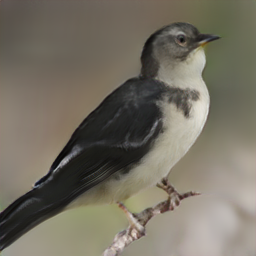}&
  \includegraphics[width=0.13\linewidth]{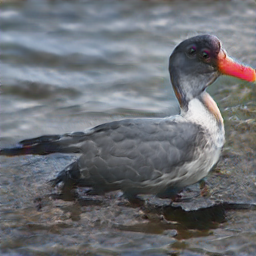}&
  \includegraphics[width=0.13\linewidth]{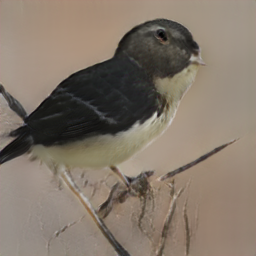}&
  \includegraphics[width=0.13\linewidth]{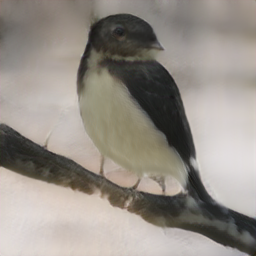}&
  \includegraphics[width=0.13\linewidth]{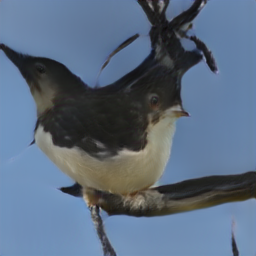}
  \\
  \multicolumn{7}{c}{this bird has wings that are {\color{red}red} and has a {\color{red}yellow} belly}
  \\
  \includegraphics[width=0.13\linewidth]{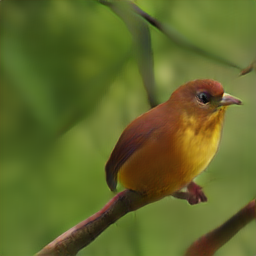}&
  \includegraphics[width=0.13\linewidth]{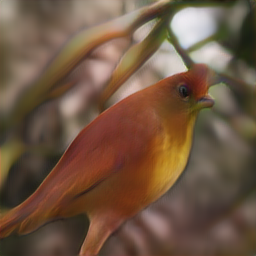}&
  \includegraphics[width=0.13\linewidth]{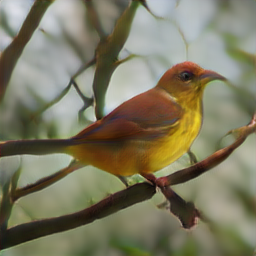}&
  \includegraphics[width=0.13\linewidth]{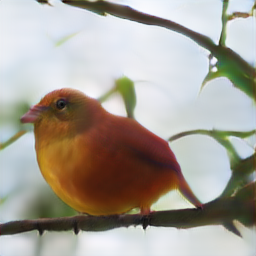}&
  \includegraphics[width=0.13\linewidth]{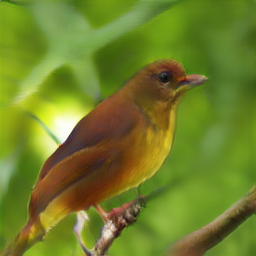}&
  \includegraphics[width=0.13\linewidth]{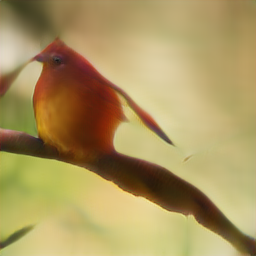}&
  \includegraphics[width=0.13\linewidth]{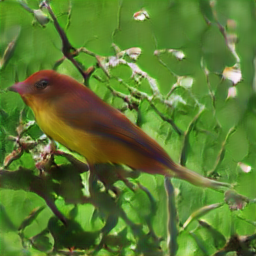}
  \\
  \multicolumn{7}{c}{this bird has wings that are {\color{red}blue} and has a {\color{red}red} belly}
  \\
  \includegraphics[width=0.13\linewidth]{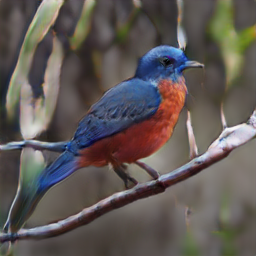}&
  \includegraphics[width=0.13\linewidth]{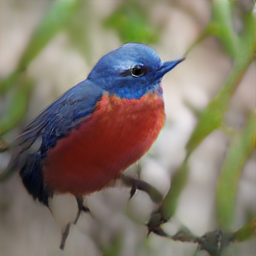}&
  \includegraphics[width=0.13\linewidth]{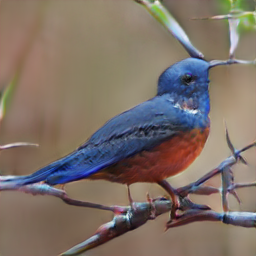}&
  \includegraphics[width=0.13\linewidth]{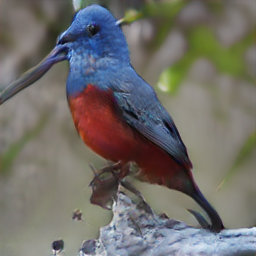}&
  \includegraphics[width=0.13\linewidth]{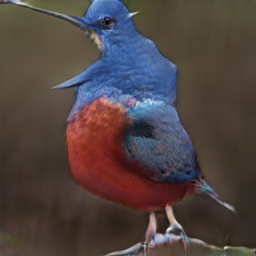}&
  \includegraphics[width=0.13\linewidth]{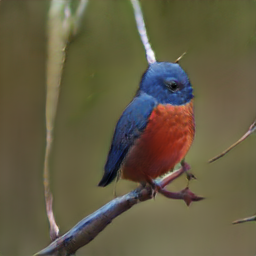}&
  \includegraphics[width=0.13\linewidth]{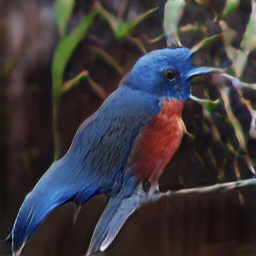}
  \\
%   \hline
 \end{tabular}
 \vspace{+3pt}
 \caption{Example results of our AttnGAN model trained on CUB while changing some most attended words in the text descriptions.}
 \vspace{-10pt}
 \label{fig:int}
 \end{figure}
\begin{figure}[h]
 \small
 \centering
 \begin{tabular}{@{\hspace{1mm}}c@{\hspace{1mm}}c@{\hspace{1mm}}c@{\hspace{1mm}}c@{\hspace{1mm}}}
%   \shortstack{\\{\color{red}a} fluffy black \\{\color{red}cat} floating on \\top {\color{red}of a lake}} 
%   & \shortstack{\\{\color{red}a red} double \\{\color{blue}decker bus} \\{\color{red}is floating} on \\top of a {\color{red}lake}} 
%   & \shortstack{\\{\color{red}a} {\color{blue}stop} {\color{red}sign \\is} floating on \\top of {\color{red}a lake}} 
%   & \shortstack{\\{\color{red}a} {\color{blue}stop} {\color{red}sign} \\is flying in \\{\color{red}the blue sky}}
  \shortstack{\\a fluffy black \\cat floating on \\top of a lake} 
  & \shortstack{\\a red double \\decker bus \\is floating on \\top of a lake} 
  & \shortstack{\\a stop sign \\is floating on \\top of a lake} 
  & \shortstack{\\a stop sign \\is flying in \\the blue sky}
  \\
  \includegraphics[width=0.245\linewidth]{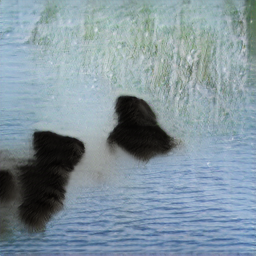} &
  \includegraphics[width=0.245\linewidth]{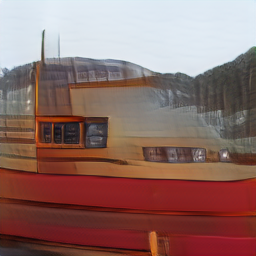} &
  \includegraphics[width=0.245\linewidth]{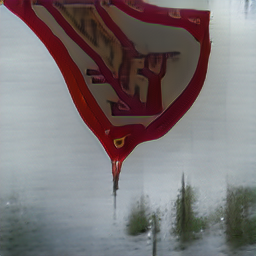} &
  \includegraphics[width=0.245\linewidth]{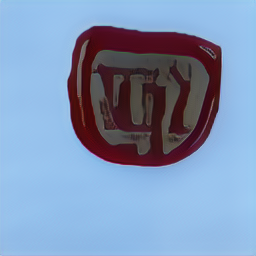}
  \\
 \end{tabular}
 \vspace{-2pt}
 \caption{256$\times$256 images generated from descriptions of novel scenarios using the AttnGAN model trained on COCO. (Intermediate results are given in the supplementary material.)}
 \vspace{-2pt}
 \label{fig:novel_coco}
 \end{figure}
\begin{figure}[h]
 \small
 \centering
 \begin{tabular}{c@{\hspace{1mm}}c@{\hspace{1mm}}c@{\hspace{1mm}}c@{\hspace{1mm}}c@{\hspace{1mm}}c@{\hspace{1mm}}c}
  \includegraphics[width=0.13\linewidth]{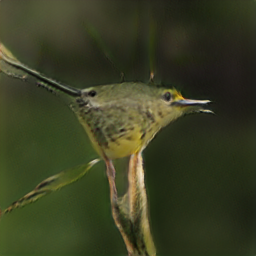}&
  \includegraphics[width=0.13\linewidth]{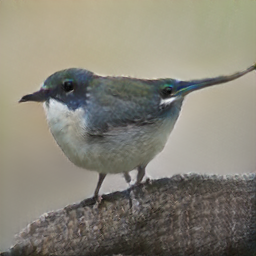}&
  \includegraphics[width=0.13\linewidth]{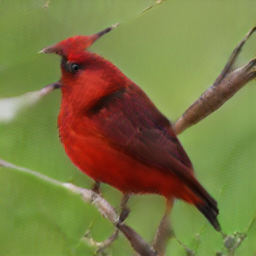}&
  \includegraphics[width=0.13\linewidth]{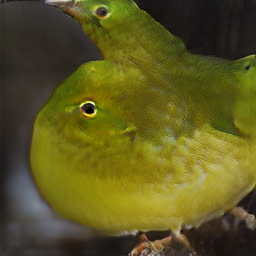}&
  \includegraphics[width=0.13\linewidth]{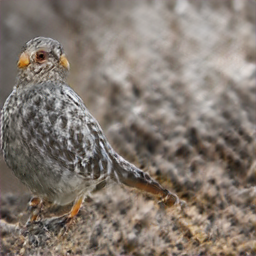}&
  \includegraphics[width=0.13\linewidth]{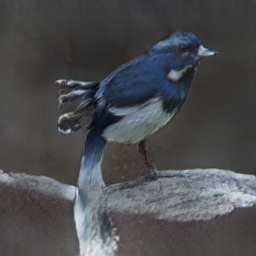}&
  \includegraphics[width=0.13\linewidth]{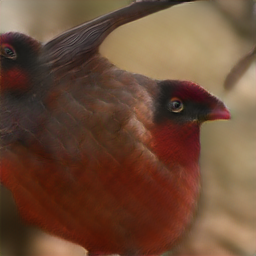}
  \\
  \includegraphics[width=0.13\linewidth]{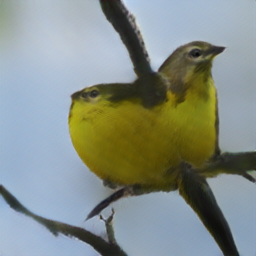}&
  \includegraphics[width=0.13\linewidth]{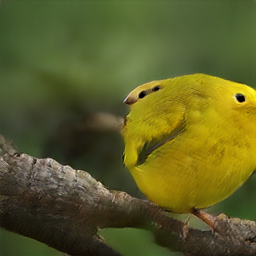}&
  \includegraphics[width=0.13\linewidth]{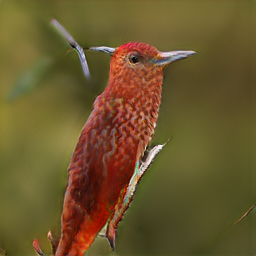}&
  \includegraphics[width=0.13\linewidth]{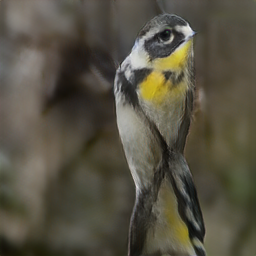}&
  \includegraphics[width=0.13\linewidth]{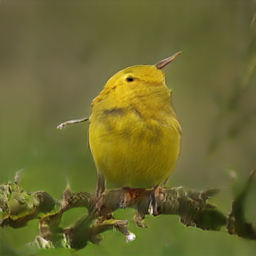}&
  \includegraphics[width=0.13\linewidth]{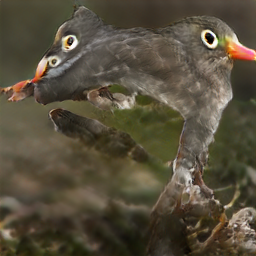}&
  \includegraphics[width=0.13\linewidth]{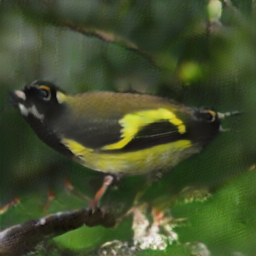}
  \\
 \end{tabular}
 \vspace{+3pt}
 \caption{Novel images by our AttnGAN on the CUB test set.}
 \vspace{-10pt}
 \label{fig:novel}
 \end{figure}
\begin{table*}[tb]
\begin{center}
% \footnotesize
\small
\begin{tabular}{l|c|c|c|c|c|c}
\hline
\multirow{1}{*}{Dataset} &GAN-INT-CLS~\cite{reed2016generative} &GAWWN~\cite{reed2016learning}   &StackGAN~\cite{Han16stackgan}  & StackGAN-v2~\cite{Han17stackgan2} & PPGN~\cite{NguyenYBDC17} & Our AttnGAN\\
\hline
CUB  &2.88~$\pm$~.04  &3.62~$\pm$~.07  &3.70~$\pm$~.04  &3.82~$\pm$~.06 &/  &\textbf{4.36~$\pm$~.03}\\
\hline
COCO  &7.88~$\pm$~.07  &/  &8.45~$\pm$~.03  &/  & 9.58~$\pm$~.21   &\textbf{25.89~$\pm$~.47}\\  
\hline
\end{tabular}
\end{center}
\vspace{-5pt}
    \caption{Inception scores by state-of-the-art GAN models~\cite{reed2016generative,reed2016learning,Han16stackgan,Han17stackgan2,NguyenYBDC17} and our AttnGAN on CUB and COCO test sets.}
\vspace{-5pt}
\label{tab:cmp_previous} 
\end{table*}

\subsection{Component analysis} \label{sec:component}
\vspace{-5pt}
In this section, we first quantitatively evaluate the AttnGAN and its variants.  The results are shown in Table~\ref{tab:component} and Figure~\ref{fig:component}. Our ``AttnGAN1'' architecture has one attention model and two generators, while the ``AttnGAN2'' architecture has two attention models stacked with three generators (see Figure~\ref{fig:architecture}). In addition, as illustrated in Figure~\ref{fig:examples}, Figure~\ref{fig:int}, Figure~\ref{fig:novel_coco}, and Figure~\ref{fig:novel}, we qualitatively examine the images generated by our AttnGAN.

\textbf{The DAMSM loss. }{
To test the proposed $\mathcal{L}_{DAMSM}$, we adjust the value of $\lambda$ (see Eq.~(\ref{eq:final-LG})).
As shown in Figure~\ref{fig:component}, a larger $\lambda$ leads to a significantly higher R-precision rate on both CUB and COCO datasets. On the CUB dataset, when the value of $\lambda$ is increased from 0.1 to 5, the inception score of the AttnGAN1 is improved from 4.19 to 4.35 and the corresponding R-precision rate is increased from 16.55\% to 58.65\% (see Table~\ref{tab:component}). On the COCO dataset, by increasing the value of $\lambda$ from 0.1 to 50, the AttnGAN1 achieves both high inception score and R-precision rate  (see Figure~\ref{fig:component}). This comparison demonstrates that properly increasing the weight of $\mathcal{L}_{DAMSM}$ helps to generate higher quality images that are better conditioned on given text descriptions. The reason is that the proposed fine-grained image-text matching loss $\mathcal{L}_{DAMSM}$ provides additional supervision (\ie, word level matching information) for training the generator. Moreover, in our experiments, we do not observe any collapsed nonsensical mode in the visualization of AttnGAN-generated images. It indicates that, with extra supervision, the fine-grained image-text matching loss also helps to stabilize the training process of the AttnGAN. In addition, if we replace the proposed DAMSM sub-network with the text encoder used in \cite{reed2016cvpr}, on the CUB dataset, the inception score and R-precision drops to 3.98 and 10.37\%, respectively (\ie, the ``AttnGAN1,~no DAMSM'' entry in table 2), which further demonstrates the effectiveness of the proposed $\mathcal{L}_{DAMSM}$. 
}

\textbf{The attentional generative network. }{
As shown in Table~\ref{tab:component} and Figure~\ref{fig:component}, stacking two attention models in the generative networks not only generates images of a higher resolution (from 128$\times$128 to 256$\times$256 resolution), but also yields higher inception scores on both CUB and COCO datasets. In order to guarantee the image quality, we find the best value of $\lambda$ for each dataset by increasing the value of $\lambda$ until the overall inception score is starting to drop on a held-out validation set. ``AttnGAN1'' models are built for searching the best $\lambda$, based on which a ``AttnGAN2'' model is built to generate higher resolution images. Due to GPU memory constraints, we did not try the AttnGAN with three attention models. As the result, our final model for CUB and COCO is ``AttnGAN2,~$\lambda$=5'' and ``AttnGAN2,~$\lambda$=50'', respectively. The final $\lambda$ of the COCO dataset turns out to be much larger than that of the CUB dataset, indicating that the proposed $\mathcal{L}_{DAMSM}$ is especially important for generating complex scenarios like those in the COCO dataset.

To better understand what has been learned by the AttnGAN, we visualize its intermediate results with attention. As shown in Figure~\ref{fig:examples}, the first stage of the AttnGAN ($G_0$) just sketches the primitive shape and colors of objects and generates low resolution images. Since only the global sentence vectors are utilized in this stage, the generated images lack details described by exact words, \eg, the beak and eyes of a bird. Based on word vectors, the following stages ($G_1$ and $G_2$) learn to rectify defects in results of the previous stage and add more details to generate higher-resolution images. Some sub-regions/pixels of $G_1$ or $G_2$ images can be inferred directly from images generated by the previous stage. For those sub-regions, the attention is equally allocated to all words and shown to be black in the attention map (see Figure~\ref{fig:examples}). For other sub-regions, which usually have semantic meaning expressed in the text description such as the attributes of objects, the attention is allocated to their most relevant words (bright regions in Figure~\ref{fig:examples}). Thus, those regions are inferred from both word-context features and previous image features of those regions.  As shown in Figure~\ref{fig:examples}, on the CUB dataset, the words \textit{the, this, bird} are usually attended by the $F^{attn}$ models for locating the object; the words describing object attributes, such as colors and parts of birds, are also attended for correcting defects and drawing details. On the COCO dataset, we have similar observations. Since there are usually more than one object in each COCO image, it is more visible that the words describing different objects are attended by different sub-regions of the image, \eg, \textit{bananas, kiwi} in the bottom-right block of Figure~\ref{fig:examples}. Those observations demonstrate that the AttnGAN learns to understand the detailed semantic meaning expressed in the text description of an image. Another observation is that our second attention model $F^{attn}_2$ is able to attend to some new words that were omitted by the first attention model $F^{attn}_1$ (see Figure~\ref{fig:examples}). It demonstrates that, to provide richer information for generating higher resolution images at latter stages of the AttnGAN, the corresponding attention models learn to recover objects and attributes omitted at previous stages. 
% , providing richer information for the attention model
}

\textbf{Generalization ability. }{
Our experimental results above have quantitatively and qualitatively shown the generalization ability of the AttnGAN by generating images from unseen text descriptions. Here we further test how sensitive the outputs are to changes in the input sentences by changing some most attended words in the text descriptions. Some examples are shown in Figure~\ref{fig:int}. It illustrates that the generated images are modified according to the changes in the input sentences, showing that the model can catch subtle semantic differences in the text description. Moreover, as shown in Figure~\ref{fig:novel_coco}, our AttnGAN can generate images to reflect the semantic meaning of descriptions of novel scenarios that are not likely to happen in the real world, \eg, \textit{a stop sign is floating on top of a lake}. 
% In sum, observations shown in these examples demonstrate the great generalization ability of the AttnGAN. 
On the other hand, we also observe that the AttnGAN sometimes generates images which are sharp and detailed, but are not likely realistic. As examples shown in Figure~\ref{fig:novel}, the AttnGAN creates birds with multiple heads, eyes or tails, which only exist in fairy tales. This indicates that our current method is still not perfect in capturing global coherent structures, which leaves room to improve. To sum up, observations shown in Figure~\ref{fig:int}, Figure~\ref{fig:novel_coco} and Figure~\ref{fig:novel} further demonstrate the generalization ability of the AttnGAN. 

% Nearest neighbor searching (Supp)
}

\subsection{Comparison with previous methods} \label{sec:compare}
\vspace{-5pt}
We compare our AttnGAN with previous state-of-the-art GAN models for text-to-image generation on CUB and COCO test sets. As shown in Table~\ref{tab:cmp_previous}, on the CUB dataset, our AttnGAN achieves 4.36 inception score, which significantly outperforms the previous best inception score of 3.82. More impressively, our AttnGAN boosts the best reported inception score on the COCO dataset from 9.58 to 25.89, a 170.25\% improvement relatively. The COCO dataset is known to be much more challenging than the CUB dataset because it consists of images with more complex scenarios. Existing methods struggle in generating realistic high-resolution images on this dataset. Examples in Figure~\ref{fig:examples} and Figure~\ref{fig:novel_coco} illustrate that our AttnGAN succeeds in generating 256$\times$256 images for various scenarios on the COCO dataset, although those generated images of the COCO dataset are not as photo-realistic as that of the CUB dataset. The experimental results show that, compared to previous state-of-the-art approaches, the AttnGAN is more effective for generating complex scenes due to its novel attention mechanism that catches fine-grained word level and sub-region level information in text-to-image generation.  

%

%-------------------------------------------------------------------------
\section{Conclusions}
\vspace{-5pt}
In this paper, we propose an Attentional Generative Adversarial Network, named AttnGAN, for fine-grained text-to-image synthesis. First, we build a novel attentional generative network for the AttnGAN to generate high quality image through a multi-stage process. Second, we propose a deep attentional multimodal similarity model to compute the fine-grained image-text matching loss for training the generator of the AttnGAN. Our AttnGAN significantly outperforms previous state-of-the-art GAN models, boosting the best reported inception score by 14.14\% on the CUB dataset and 170.25\% on the more challenging COCO dataset. Extensive experimental results clearly demonstrate the effectiveness of the proposed attention mechanism in the AttnGAN, which is especially critical for text-to-image generation for complex scenes.

{\footnotesize
% {\small
\bibliographystyle{ieee}
\bibliography{paperbib}
}

% \clearpage
% \onecolumn
\section*{Supplementary Material} 
Due to the size limit, more examples are available in the appendix, which can be  download from {\href{https://1drv.ms/b/s!Aj4exx_cRA4ghK5-kUG-EqH7hgknUA} {this link}}.

\end{document}